\documentclass[letterpaper, 10 pt, conference]{ieeeconf}  

\IEEEoverridecommandlockouts                              

\usepackage{amsmath}
\usepackage{amssymb}
\usepackage{multirow}
\usepackage{booktabs}
\usepackage{array}
\usepackage{subfigure}
\usepackage{graphicx}
\usepackage{cite}
\usepackage{makecell}
\usepackage{color}
\usepackage[capitalize]{cleveref}

\title{\LARGE \bf
FlowDepth: Decoupling Optical Flow for Self-Supervised Monocular Depth Estimation
}

\author{Yiyang Sun, Zhiyuan Xu, Xiaonian Wang and Jing Yao$^{\dag}$
\thanks{$^{\dag}$indicates the corresponding author.}
\thanks{Y.Sun, Z.Xu, X.Wang, and J.Yao are with the College of Electronical and Information Engineering, Tongji University, China. E-mail: {\tt\small{\{2130734, zhiyuan, dawnyear, yaojing\}@tongji.edu.cn}}}
}

\begin{document}

\maketitle
\thispagestyle{empty}
\pagestyle{empty}

\begin{abstract}

Self-supervised multi-frame methods have currently achieved promising results in depth estimation. However, these methods often suffer from mismatch problems due to the moving objects, which break the static assumption. Additionally, unfairness can occur when calculating photometric errors in high-freq or low-texture regions of the images. To address these issues, existing approaches use additional semantic priori black-box networks to separate moving objects and improve the model only at the loss level. Therefore, we propose FlowDepth, where a Dynamic Motion Flow Module (DMFM) decouples the optical flow by a mechanism-based approach and warps the dynamic regions thus solving the mismatch problem. For the unfairness of photometric errors caused by high-freq and low-texture regions, we use Depth-Cue-Aware Blur (DCABlur) and Cost-Volume sparsity loss respectively at the input and the loss level to solve the problem. Experimental results on the KITTI and Cityscapes datasets show that our method outperforms the state-of-the-art methods.

\end{abstract}


\section{Introduction}
\label{sec:intro}

    \begin{figure*}
        \centering
        \vspace*{0.25cm}
        \includegraphics[width=0.9\textwidth, trim={0 6 0 0}]{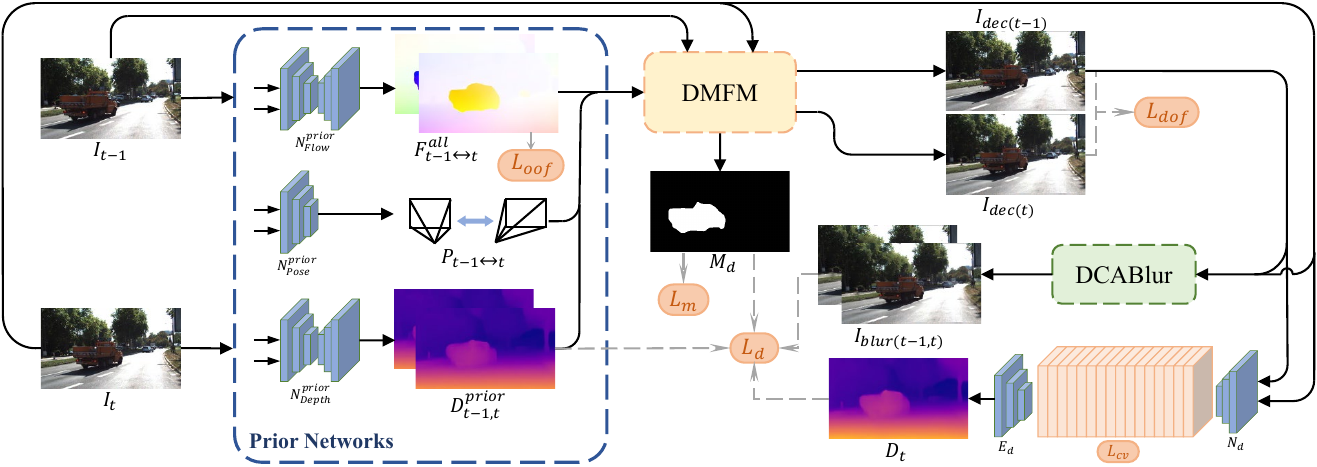}
        \caption{Architecture of our FlowDepth. The images $I_{t-1}$ and $I_t$ are first passed through the prior networks to get depth, camera motion, and optical flow prior. Then DMFM decouples the moving objects with these prior. The new images are fed into a multi-frame depth estimation network constrained by the cost-volume sparse loss. Finally, it generates the depth estimation results. Before calculating the multi-frame depth reprojection loss, the images will go through the DCABlur module for blurring to mitigate high-freq texture problems.}
        \label{fig:model}
        \vspace{-0.6cm}
    \end{figure*}

    Estimating the depth of each pixel in an image is an essential task for obtaining 3D scene geometry information, which provides necessary geometric cues in robot navigation and self-driving. However, depth labels for supervised training are always difficult to obtain, and even using LiDAR sensors to capture depth ground truth can only provide sparse points compared to the density of pixels. Since spatiotemporal continuous images are commonly available in reality, the self-supervised way \cite{ Zhou_Brown_Snavely_Lowe_2017, Watson_Mac_Aodha_Prisacariu_Brostow_Firman_2021, Feng_Yang_Jing_Wang_Tian_Li, Chen_Zhang_Jiang_Wang_Li_Li_2022}, which minimizes the photometric error between the original and synthetic images, has become more popular in recent years.

    Early convolutional self-supervised depth estimation methods \cite{Godard_Aodha_Brostow_2016, Godard_Aodha_Firman_Brostow_2018} only use adjacent frames to compute reprojection loss in training, but they do not utilize geometric constraint information of the temporal sequence frames in inference, which limits the performance of depth estimation. Therefore, recent methods \cite{Watson_Mac_Aodha_Prisacariu_Brostow_Firman_2021, Wimbauer_Yang_Stumberg_Zeller_Cremers_2020} use multi-frame to construct a similar cost-volume in stereo-match tasks during inference to improve estimation accuracy.

    However, when using reprojection loss to calculate the photometric error, there are two problems that break the photometric consistency assumption. In \textbf{low-texture regions} such as road surfaces or building wall surfaces, a low photometric error will be calculated even if the matched pixel has a large deviation from $gt$ \cite{Jiao_Tran_Shi_2020}. In \textbf{high-freq texture regions} such as leaves or the edges of pavement lanes, a small deviation from $gt$ may result in a large photometric error \cite{Chen_Li_Zhang_Li_2022}. In addition, both the reprojection loss and the cost-volume-based approach suffer from the \textbf{dynamic objects} which can violate the static environment assumption and lead to the mismatch problem in the dynamic regions. Recent works usually introduce auxiliary semantic priors to mask out dynamic objects to solve this problem. Nevertheless, the accuracy is directly influenced by the pre-trained semantic network.

    Therefore, we propose FlowDepth in order to solve these problems and improve the depth estimation accuracy. First, we predict depth, camera motion, and optical flow prior, and decouple the optical flow into a static/rigid part (caused by ego camera motion) and a dynamic part (caused by dynamic object motion) through Dynamic Motion Flow Module (DMFM). Then the dynamic part is applied to the source frame and 'moves' the dynamic object to where it should be if it keeps the global position in the target frame. The new source frame and target frame can guarantee the static environment assumption. Then, we use the Depth-Cue-Aware Blur (DCABlur) module to only blur the high-freq regions or edges with dramatic color changes to increase the one-pixel perceptual field to make the photometric error fairer, while ensuring the depth cues in the image are not affected by blur. In the training phase, we propose cost-volume sparse loss to alleviate the unfairness of reprojection loss in low-texture regions. We summarize the contributions of our paper as follows:

    \begin{itemize}
        \item We propose a novel Dynamic Motion Flow Module (DMFM) that decouples the optical flow by a mechanism-based approach and then warps the dynamic objects in the source frame to solve the mismatch problem with no additional annotations.
        \item We design a Depth-Cue-Aware Blur (DCABlur) module and a cost-volume sparse loss to mitigate the reprojection mismatch problem.
        \item We experimentally validate the performance of our method on KITTI, Cityscapes, and our own VECAN datasets. The results show that our model outperforms prior methods.
    \end{itemize}

\section{Related Works}
\label{sec:relatedworks}

\subsection{Single-frame monocular depth}
    ~\cite{Zhou_Brown_Snavely_Lowe_2017} first proposes a self-supervised depth estimation framework that simultaneously predicts depth and ego-motion. Based on \cite{Zhou_Brown_Snavely_Lowe_2017}, Monodepth2 \cite{Godard_Aodha_Firman_Brostow_2018} introduces per-pixel minimization reprojection loss and auto-masking to address the dynamic objects problem. In order to meet the real-time requirements, PackNet \cite{Guizilini_Ambrus_Pillai_Raventos_Gaidon_2019} utilizes 3D convolution to explore more effective model. Lite-Mono \cite{Zhang_Nex_Vosselman_Kerle_2022} further designs a lightweight model based on the transformer and CNN. RM-depth \cite{Hui} tackles the issue from the perspective of model volume, employing RMU to substantially reduce the model's parameters. 
    The idea of all these methods is to project adjacent frames during training to satisfy reprojection photometric consistency \cite{Wang_Bovik_Sheikh_Simoncelli_2004}. However, the reprojection loss will meet the mismatch problem caused by dynamic objects or the high-freq/low-texture region. In contrast, our proposed DMFM, DCABlur module, and CV Loss can solve the mismatch problem, thereby improving the accuracy of depth estimation.

\subsection{Multi-frame monocular depth}
    Due to the aforementioned methods using only a single image during inference, it fails to consider temporal constraints, thereby limiting the model's performance. Therefore, a direct approach is to incorporate multi-frame inputs by utilizing recurrent networks both during the training and inference stages. \cite{Patil_Gansbeke_Dai_Gool_2020} employed ConvLSTM to predict the depth sequentially. Compared to recurrent networks, MonoRec \cite{Wimbauer_Yang_Stumberg_Zeller_Cremers_2020} and Manydepth \cite{Watson_Mac_Aodha_Prisacariu_Brostow_Firman_2021} introduce the cost-volume in stereo-match, enabling geometric constraints during inference and greatly saving the inference time. However, cost-volume-based methods still rely on static environment assumptions. In contrast, our DMFM module fundamentally 'staticizes' all dynamic objects at the input level, thereby enhancing the depth estimation in dynamic object regions.

\subsection{Depth estimation with auxiliary tasks}

    \begin{figure*}
        \vspace*{0.25cm}
        \centering
        \setlength{\abovecaptionskip}{0.1cm}
        \includegraphics[width=0.85\textwidth]{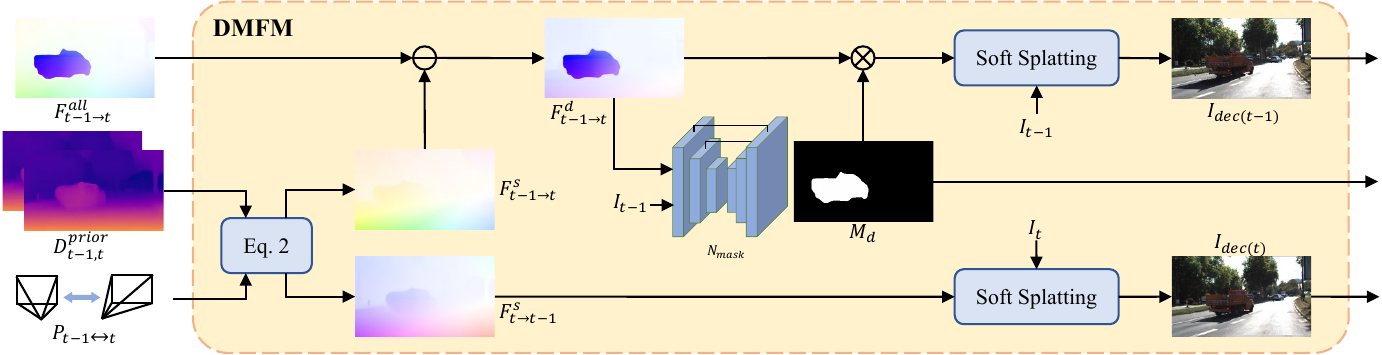}
        \caption{The detailed structure of DMFM. Firstly, the static optical flow $F^s$ is obtained using depth and pose priors. Then, the overall optical flow $F^{all}_{t-1 \rightarrow t}$ is decoupled by $F^s_{t-1 \rightarrow t}$ to obtain the dynamic optical flow $F^d_{t-1 \rightarrow t}$. By learning a mask network, $I_{t-1}$ is warped according to the $F^d_{t-1 \rightarrow t}$ in dynamic area to get $I_{dec(t-1)}$. Meanwhile, by directly applying the $F^s_{t \rightarrow t-1}$ to $I_{t}$, we can also obtain $I_{dec(t)}$. Theoretically, $I_{dec(t-1)}$ and $I_{dec(t)}$ are the same.}
        \label{fig:dmfm}
        \vspace{-0.6cm}
    \end{figure*}

    
    Prior works address the issue of dynamic objects by employing object motion prediction networks  or additional semantic prior. Since both the depth estimation and the flow estimation adhere to the scene geometric consistency constraint, \cite{Zou_Luo_Huang_2018, Ranjan_Jampani_Balles_Kim_Sun_Wulff_Black_2019, Chen_Schmid_Sminchisescu_2019} simultaneously predict depth and optical flow using a multitask model to enhance the results. Nevertheless, multitask models are usually hard to converge during the training. \cite{Casser_Pirk_Mahjourian_Angelova_2018, Lee_Im_Lin_Kweon_2021, Jiao_Tran_Shi_2020, Lee_Rameau_Pan_Kweon_2021, Gordon_Li_Jonschkowski_Angelova_2020} try to predict the dynamic objects motion using black-box networks, followed by pixel-wise warping of the dynamic object.  DynamicDepth \cite{Feng_Yang_Jing_Wang_Tian_Li} directly uses a pre-trained segmentation network to separate objects and performs warping with depth prior. TriDepth \cite{Chen_Zhang_Jiang_Wang_Li_Li_2022} even directly uses the semantic labels to get object edges and alleviate the ubiquitous edge-fattening issue by using triplet loss. However, these approaches have drawbacks such as the low interpretability of the black-box object motion network, and the need for additional segmentation networks that should be trained with expensive manual annotations. In contrast, our proposed DMFM decouples the optical flow using a mechanism-based approach, which is consistent with the scene geometry. So the whole training process requires only image data without any manual annotations.

\section{Methodology}
\label{sec:method}


\subsection{Overall Structure}
\label{sec:structure}
    
    As shown in Fig.\ref{fig:model}, We first use prior networks to predict depth, camera motion, and optical flow prior, which are sent to the Dynamic Motion Flow Module (Sec.\ref{sec:dmfm}) to decouple the moving objects in $I_{t-1}$ in order to solve the mismatch problem caused by dynamic objects. Then the decoupled frame $I_{dec(t-1)}$ and $I_t$ are Depth-Cue-Aware Blurred (Sec.\ref{sec:dcab}) in the high-freq regions in images to mitigate the unfairness of reprojection loss. Finally, they are sent to a cost-volume-based depth estimation network to predict the depth. To solve the depth uncertainty caused by unfair photometric errors in the low-texture regions, we introduce the cost-volume sparse loss (Sec.\ref{sec:loss}).
    
\subsection{Dynamic Motion Flow Module (DMFM)}
\label{sec:dmfm}
    \textbf{Preliminaries.} When there is relative motion between the camera and the scene, the optical features of the scene object surface are projected onto the image plane, resulting in the optical flow. Given two continuous frames $I_{t-1}$ and $I_t$, the pixel motion (denoted as flow $F^{all}$) can be decomposed into two components: one is the static/rigid flow $F^s$ representing the background variations in the scene caused by the ego-motion, and the other is the dynamic flow $F^d$ induced by the self-motion of moving objects in the scene. The $F^s$ adheres to the geometric consistency of the scene and is related to the scene depth and camera motion. When a 2D point $\rm \textbf{x} \in\mathbb{R}^2$ in the image is given, it can be reprojected back to $\rm \textbf{X} \in\mathbb{R}^3$ in the 3D world coordinate system:
        \begin{equation}
        \label{eq:project}
            {\rm \textbf{X}} ^ {\rm T} = d({\rm \textbf{x}}) K ^ {-1} [\rm \textbf{x};1] ^ {\rm T},
        \end{equation}
    where $K\in\mathbb{R}^{3\times3}$ is the camera intrinsic parameter, $d$ represents the depth in 3D world corresponding to each pixel. Based on Eq. \ref{eq:project}, we can calculate the difference between ${\rm \textbf{x}}_{t-1}$ and ${\rm \textbf{x}}_{t}$, which represents the static flow $F^s$:
        \begin{equation}
        \label{eq:cal_static_flow}
            F^{s}_{t-1 \rightarrow t} = \frac{1}{d_{t}}KP(d_{t-1}({\rm\textbf{x}}_{t-1}) K ^ {-1} [{\rm \textbf{x}}_{t-1};1]^{\rm T})) - [{\rm \textbf{x}}_{t-1};1]^{\rm T},
        \end{equation}
    where the first term calculates as ${\rm \textbf{x}}_{t}$, $P = [R;T]\in\mathbb{R}^{3\times4}$ is the relative pose change of the camera, and $d_{t}$ is the normalization coefficient form 3D to 2D, which is commonly cosidered as a constant. In Eq. \ref{eq:cal_static_flow}, once we have $d_{t-1}$ and $P$, we can obtain $F^{s}_{t-1 \rightarrow t}$, which represents the static flow from $t-1$ to $t$.
    
    \begin{figure*}[htbp]
        \vspace*{0.25cm}
        \centering
        \setlength{\abovecaptionskip}{0.1cm}
        \includegraphics[width=0.85\textwidth]{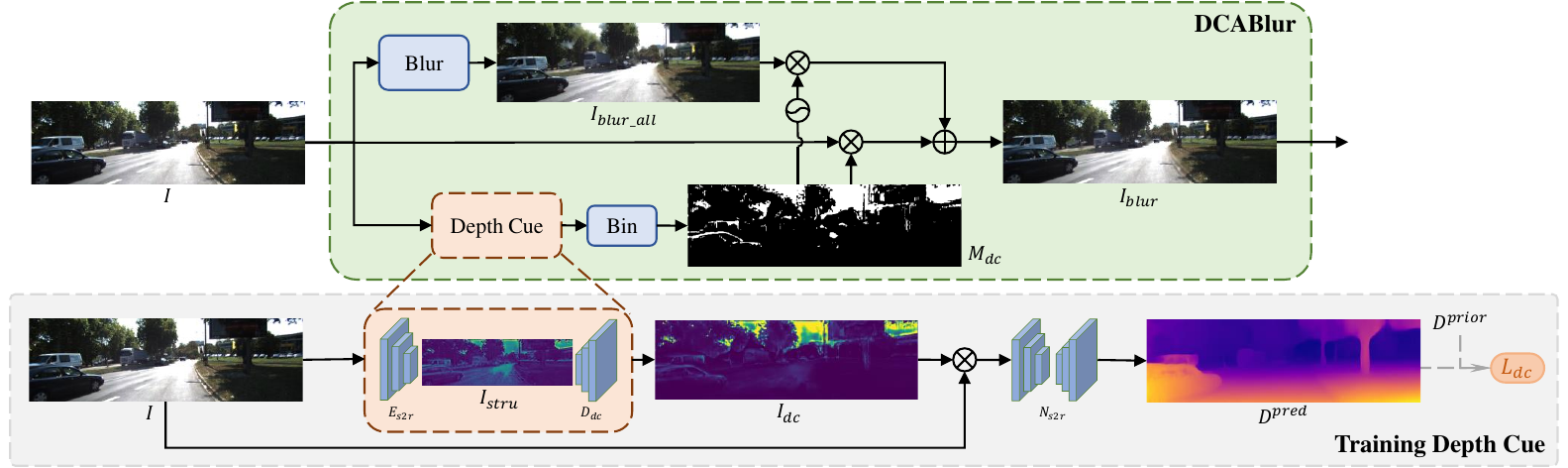}
        \caption{Illustration of DCABlur. The Depth Cue is pre-trained as shown in the grey area.}
        \label{fig:dcab}
        \vspace{-0.6cm}
    \end{figure*}
    
    \textbf{DMFM implementation.}
    The key idea of DMFM is to relocate the moving objects in the source frame ($I_{t-1}$) to where they should be if the objects are stationary in the target frame ($I_{t}$). In other words, it can be seen as fixing the dynamic objects in $t-1$ at their world coordinates in $t$. We denote the new frame after warping the objects as $I_{dec}$.
    
    As shown in Fig.\ref{fig:dmfm}, we first obtain depth, pose, and optical flow (all) priors through the teacher networks. As mentioned in the preliminaries, we can decompose the overall flow between two frames into dynamic and static flow components. Therefore, we have two approaches to obtain $I_{dec}$ from $I_{t-1}$ and $I_{t}$ respectively. The first one is to obtain the dynamic motion flow $F^{d}_{t-1 \rightarrow t}$ by
        \begin{equation}
        \label{eq:fd_t-1}
            F^{d}_{t-1 \rightarrow t} = F^{all}_{t-1 \rightarrow t} - F^{s}_{t-1 \rightarrow t}.
        \end{equation}
    $F^{all}_{t-1 \rightarrow t}$ is the optical flow prior obtained from RAFT \cite{Teed_Deng_2020}, and $F^{s}_{t-1 \rightarrow t}$ is computed using the depth and pose prior in Eq.\ref{eq:cal_static_flow}. Then $I_{dec(t-1)}$ is obtained by applying forward warping on $I_{t-1}$ using $F^{d}_{t-1 \rightarrow t}$. For the second approach, we only compute $F^{s}_{t \rightarrow t-1}$ using Eq.\ref{eq:cal_static_flow} and forward warp $I_t$ by $F^{s}_{t \rightarrow t-1}$ to get $I_{dec(t)}$. In theory, $I_{dec(t-1)}$ and $I_{dec(t)}$ should be almost the same. Finally, these two approaches are simultaneously employed, and $I_{dec(t-1)}$ is selected as the lookup frame for multi-frame depth estimation. 
    
    \textit{\textbf{Why both approaches are used?}} Although $I_{dec(t-1)}$ and $I_{dec(t)}$ should ideally be consistent, due to the imperfect estimation from the teacher networks, they may be slightly different. Therefore, we introduce an additional self-supervised loss called $L_{dof}$, which will be described in detail in the Appendix. The introduction of this loss not only improves the performance of the teacher network but also enhances the quality of $I_{dec}$, thus improving the performance of multi-frame depth estimation. Additionally, when the dynamic objects in $I_{t-1}$ are warped, part of pixels at the original location of the objects will have no value due to the occlusion. These pixels can be filled in by the background information, which typically does not have foreground occlusions in $I_{dec(t)}$, effectively resolving the occlusion issue.
    
    \textit{\textbf{Why don't we use $I_{dec(t)}$ as the lookup frame?}} This is because $I_{dec(t)}$ is obtained by warping $I_{t}$ with $F^{s}_{t \rightarrow t-1}$, which heavily relies on the depth and pose priors of $I_{t}$ and totally loses the information of $I_{t-1}$. So the inaccurate estimation results from the teacher network will limit the performance of the multi-frame network.
    
    Then, to achieve better warp results, we adopt softmax splatting \cite{Niklaus_Liu_2020} for forward warping. This approach effectively resolves the issue of multi-to-one mapping in forward warping and allows for gradient computation.

    Finally, since the inaccurate prior depth estimate makes it difficult to subtract $F^{all}_{t-1 \rightarrow t}$ and $F^{s}_{t-1 \rightarrow t}$ to equal zero in the background region, we use a U-Net, denoted as $N_{mask}$, to obtain a valid dynamic motion flow region:
        \begin{equation}
        \label{eq:get mask}
            M_d = [\theta(N_{mask}(F^{d}_{t-1 \rightarrow t}, I_{t-1}))>0.6],
        \end{equation}
    where $\theta$ is the Sigmoid function and $[\cdot]$ is the Iverson bracket. $M_d=1$ means that there is a moving object and we only do soft splatting here.
    
\subsection{Depth-Cue-Aware Blur (DCABlur)}
\label{sec:dcab}

    
    The reprojection loss used in self-supervised depth estimation calculates photometric error to indirectly constrain the depth information. Consequently, it can lead to a large loss for small depth errors due to drastic color changes in high-frequency regions, or a small loss for large depth errors due to the same color in low-texture regions. These situations result in unfairness in calculating photometric errors. Therefore, the concept of auto-blur \cite{Chen_Li_Zhang_Li_2022} is to compute a frequency map by calculating the color differences
    and then the regions with frequencies above a threshold are blurred.
    
    We propose the DCABlur module based on auto-blur as shown in Fig.\ref{fig:dcab}. The high-freq edges obtained by auto-blur include both texture edges and depth edges. Texture edges often correspond to small depth variations and are suitable for blurring methods. On the other hand, depth edges typically have significant depth changes, and blurring them would reduce the overall photometric error, resulting in a worse distinction between foreground and background. Therefore, our DCABlur aims to identify depth edges in images and only applies blurring to texture edges.

    Inspired by \cite{Chen_Wang_Chen_Zeng_2021}, we found that depth cues are inherent semantic properties of objects, which do not vary with the style and texture of the object. Whereas a scene only consists of style information and structure information, we infer that depth cues are only related to structure information. So we pre-train a network to extract depth cues useful for depth estimation in images as shown in Training Depth Cue of Fig.\ref{fig:dcab}. The network is primarily an encoder-decoder model, where the structure encoder ($E_{s2r}$) uses the pre-trained style transfer encoder in \cite{Chen_Wang_Chen_Zeng_2021} and the depth cue decoder ($D_{dc}$) is randomly initialized. The depth estimation network ($N_{s2r}$) also uses the pre-trained one in \cite{Chen_Wang_Chen_Zeng_2021} to output a depth map for supervised learning. In training, we fix the structure encoder and the depth estimation network and only train the depth cue decoder to find the depth cue. $D^{pior}$ is directly used as the ground truth. Because we are not looking for a particularly good depth output from this network, we just need the decoder to be able to learn the depth cues that the network considers helpful for depth estimation. The loss function is:
        \begin{equation}
        \label{eq:depth cue loss}
        \begin{aligned}
            L_{dc} &= \frac{1}{HW}\Vert N_{s2r}(D_{dc}(E_{s2r}(I)) \otimes I) - D^{prior} \Vert_1 \\
            &+ \frac{\sigma}{HW} \Vert D_{dc}(E_{s2r}(I)) \Vert_1,
        \end{aligned}
        \end{equation}
    where the first term is directly supervised using an L1 loss and the second term is used to enforce the sparsity of the depth cue mask. $H$, $W$ are the height and width of input image, $\sigma$ is a coefficient which controls the degree of sparsity.
    
    Finally, the trained depth cue network ($E_{s2r}$+$D_{dc}$) is directly integrated into the DCABlur module and then the blur operation is applied to the image with the depth cue mask $M_{dc}$.

\subsection{Loss Functions}
\label{sec:loss}

    When training our FlowDepth framework, we introduce a new cost-volume loss function to address the unfair photometric error in low-texture regions. Additionally, since we propose several modules, corresponding loss functions are designed for these modules as well. 

\begin{figure}[tb]
        \vspace{0.2cm}
        \centering
        \setlength{\abovecaptionskip}{0.cm}
        \includegraphics[width=\linewidth]{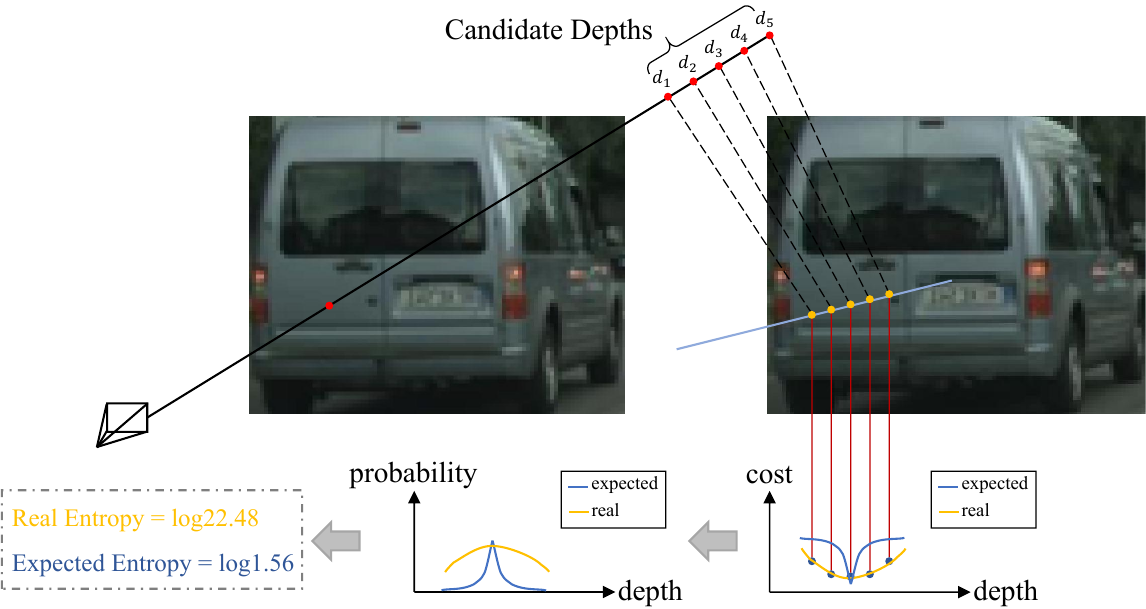}
        \caption{The depth uncertainty caused by low-texture regions in cost volume. When the pixels in the low-texture region are projected to the matching map according to the candidate depths, several extremely close losses are obtained. Ideally, the probability of the feature cost in the candidate depth domain should show an unimodal distribution, but the real entropy is much larger than the expected one.}
        \label{fig:cvloss}
        \vspace{-0.6cm}
    \end{figure}

    \textbf{Cost-Volume Sparse Loss.} For more information about cost volume please refer to \cite{Watson_Mac_Aodha_Prisacariu_Brostow_Firman_2021}. In short, each channel of the cost volume represents equally spaced candidate depths. Therefore, for each pixel in the image, we aim to have the minimum matching cost at the true depth. However, as shown in Fig.\ref{fig:cvloss}, due to the low-texture regions, pixels in this area may get multiple small matching costs in the cost volume. To address this issue, we introduce probabilities as constraints, where smaller matching costs correspond to higher probability values. The depth estimation for each pixel is approximated as a classification task, and its corresponding entropy can be computed as:
        \begin{equation}
        \begin{aligned}
        \label{eq:entropy}
            En(d) &= -\sum_{i=1}^n P(d=d_i)\log P(d=d_i) \\
            & \leq  -\sum_{i=1}^k \frac{1}{k} \log (\frac{1}{k}) = \log k,
        \end{aligned}
        \end{equation}
    where $n$ is the number of candidate depths, and the entropy is maximized when the random variable follows a uniform distribution. In other words, we can control the number of confident indices with $k$ by an entropy boundary loss
        \begin{equation}
        \label{eq:entropy boundary loss}
            L_{entropy} = \frac{1}{HW} \sum_{{\rm \textbf{x}}\in I} \max (0, En_{\rm \textbf{x}} -\log k).
        \end{equation}
    Since the number of candidate depths is usually large ($n=90$) and the desired number of confident indices is small ($k=3$), we apply $L_{1/2}$ sparsity loss in order to encourage more sparse constraints on the probabilities:
        \begin{equation}
        \label{eq:sparsity loss}
            L_{sparsity} = 2 \overline{P(d)} \sum_{i=n}^n \sqrt{1 + \frac{P(d=d_i)}{\overline{P(d)}}}.
        \end{equation}
    The final cost-volume loss is
        \begin{equation}
        \label{eq:cv loss}
            L_{cv} = L_{entropy} + L_{sparsity}.
        \end{equation}

    Besides, we use and design several other self-supervised loss functions. In brief, \textbf{Depth Loss} $L_{d}$ is to train the final multi-frame depth estimation. \textbf{Origin Optical Flow Loss} $L_{oof}$ is to constrain the outputs of $N_{Flow}^{prior}$. \textbf{Decouple Optical Flow Loss} $L_{dof}$ is to get a better quality decoupled lookup frame.\textbf{ Mask Loss} $L_m$ is used to train $U_{mask}$.

    Finally, all losses are combined to train FlowDepth as:
        \begin{equation}
        \label{eq:all loss}
            L_{m} = \lambda_{cv}L_{cv} + \lambda_{d}L_{d} + \lambda_{oof}L_{oof} + \lambda_{dof}L_{dof} + \lambda_{m}L_{m}.
        \end{equation}
    where $\lambda_{cv/d/oof/dof/m}$ are learnable parameters and are used to balance the loss items.

\section{Experiments}
\label{sec:exp}

\subsection{Implementation Details}

    \begin{table*}
        \caption{Comparison on the KITTI and Cityscapes Datasets. The best in each metric is in \textbf{BOLD} and the second best is \underline{UNDERLINED}. \\ * means that the model directly uses the semantic ground truth as auxiliary information. ADDTL means whether the model requires additional manual labeling for self-supervised training across different datasets}
        \vspace{-10pt}
        \label{tab:results}
        \begin{center}
            \resizebox{\textwidth}{!}{
            \begin{tabular}{c|lccc|cccc|ccc}
                \toprule[0.5mm]
                & Method & Frames & ADDTL & W$\times$H & AbsRel$\downarrow$ & SqRel$\downarrow$ & RMSE$\downarrow$ & RMSE log$\downarrow$ & $\delta<1.25 \uparrow$ & $\delta<1.25^2 \uparrow$ & $\delta<1.25^3 \uparrow$ \\
                \midrule[0.5mm]
                \multirow{15}{*}{\rotatebox{90}{KITTI}} & Zhou et al. \cite{Zhou_Brown_Snavely_Lowe_2017} & 1 & & 416$\times$128 & 0.208 & 1.768 & 6.856 & 0.283 & 0.678 & 0.885 & 0.957 \\
                & Struct2depth \cite{Casser_Pirk_Mahjourian_Angelova_2018} & 1 & $\bullet$ & 416$\times$128 & 0.141 & 1.026 & 5.291 & 0.215 & 0.816 & 0.945 & 0.979 \\
                & CC \cite{Ranjan_Jampani_Balles_Kim_Sun_Wulff_Black_2019} & 1 & & 832$\times$256 & 0.140 & 1.070 & 5.326 & 0.217 & 0.826 & 0.941 & 0.975 \\
                & GLNet \cite{Chen_Schmid_Sminchisescu_2019} & 1 & & 416$\times$128 & 0.135 & 1.070 & 5.230 & 0.210 & 0.841 & 0.948 & 0.980 \\
                & Gordon et al. \cite{Gordon_Li_Jonschkowski_Angelova_2020} & 1 & $\bullet$ & 416$\times$128 & 0.128 & 0.959 & 5.230 & 0.212 & 0.845 & 0.947 & 0.976 \\
                & Monodepth2 \cite{Godard_Aodha_Firman_Brostow_2018} & 1 & & 640$\times$192 & 0.115 & 0.903 & 4.863 & 0.193 & 0.877 & 0.959 & 0.981\\
                & Packnet-SFM \cite{Guizilini_Ambrus_Pillai_Raventos_Gaidon_2019} & 1 & & 640$\times$192 & 0.111 & 0.785 & 4.601 & 0.189 & 0.878 & 0.960 & 0.982\\
                & RM-Depth \cite{Hui} & 1 & & 640$\times$192 & 0.107 & \underline{0.687} & 4.476 & 0.181 & 0.883 & 0.964 & \textbf{0.984}\\
                & Lite-Mono \cite{Zhang_Nex_Vosselman_Kerle_2022} & 1 & & 640$\times$192 & 0.107 & 0.765 & 4.561 & 0.183 & 0.886 & 0.963 & \underline{0.983}\\
                & FSRE-Depth \cite{Jung_Park_Yoo_2021} & 1 & $\bullet$ & 640$\times$192 & 0.105 & 0.722 & 4.547 & 0.182 & 0.886 & 0.964 & \textbf{0.984}\\
                & Patil et al. \cite{Patil_Gansbeke_Dai_Gool_2020} & N & & 640$\times$192 & 0.111 & 0.821 & 4.650 & 0.187 & 0.883 & 0.961 & 0.982\\
                & ManyDepth \cite{Watson_Mac_Aodha_Prisacariu_Brostow_Firman_2021} & 2 (-1,0) & & 640$\times$192 & 0.098 & 0.770 & 4.459 & 0.176 & \underline{0.900} & \underline{0.965} & \underline{0.983}\\
                & DynamicDepth \cite{Feng_Yang_Jing_Wang_Tian_Li} & 2 (-1,0) & $\bullet$ & 640$\times$192 & \underline{0.096} & 0.720 & \underline{4.458} & \underline{0.175} & 0.897 & \underline{0.965} & \textbf{0.984}\\
                & \textbf{FlowDepth(ours)} & 2 (-1,0) & & 640$\times$192 & \textbf{0.093} & \textbf{0.681} & \textbf{4.232} & \textbf{0.172} & \textbf{0.904} & \textbf{0.966} & \textbf{0.984}\\
                \cline{2-12}
                \rule{0pt}{9pt}
                & TriDepth$^{*}$ \cite{Chen_Zhang_Jiang_Wang_Li_Li_2022} & 2 (-1,0) & $\bullet$ & 640$\times$192 & \underline{0.093} & \textbf{0.665} & \underline{4.272} & \underline{0.172} & \textbf{0.907} & \textbf{0.967} & \textbf{0.984}\\
                & \textbf{FlowDepth$^{*}$(ours)} & 2 (-1,0) & $\bullet$ & 640$\times$192 & \textbf{0.091} & \underline{0.675} & \textbf{4.243} & \textbf{0.170} & \underline{0.905} & \textbf{0.967} & \textbf{0.984}\\
                \hline\hline
                \rule{0pt}{11pt}
                \multirow{9}{*}{\rotatebox{90}{Cityscapes}} & Struct2depth \cite{Casser_Pirk_Mahjourian_Angelova_2018} & 1 & $\bullet$ & 416$\times$128 & 0.145 & 1.737 & 7.280 & 0.205 & 0.813 & 0.942 & 0.976\\
                & Monodepth2 \cite{Godard_Aodha_Firman_Brostow_2018} & 1 & & 416$\times$128 & 0.129 & 1.569 & 6.876 & 0.187 & 0.849 & 0.957 & 0.983\\
                & Gordon et al. \cite{Gordon_Li_Jonschkowski_Angelova_2020} & 1 & $\bullet$ & 416$\times$128 & 0.127 & 1.330 & 6.960 & 0.195 & 0.830 & 0.947 & 0.981\\
                & Li et al. \cite{Li_Gordon_Zhao_Casser_Angelova_2020} & 1 & & 416$\times$128 & 0.119 & 1.290 & 6.980 & 0.190 & 0.846 & 0.952 & 0.982\\
                & Lee et al. \cite{Lee_Rameau_Pan_Kweon_2021} & 1 & $\bullet$ & 832$\times$256 & 0.116 & 1.213 & 6.695 & 0.186 & 0.852 & 0.951 & 0.982\\
                & InstaDM \cite{Lee_Im_Lin_Kweon_2021} & 1 & $\bullet$ & 832$\times$256 & 0.111 & 1.158 & 6.437 & 0.182 & 0.868 & 0.961 & 0.983\\
                & ManyDepth \cite{Watson_Mac_Aodha_Prisacariu_Brostow_Firman_2021} & 2 (-1,0) & & 512$\times$192 & 0.114 & 1.193 & 6.223 & 0.170 & 0.875 & 0.967 & \underline{0.989}\\
                & DynamicDepth \cite{Feng_Yang_Jing_Wang_Tian_Li} & 2 (-1,0) & $\bullet$ & 512$\times$192 & \underline{0.103} & \underline{1.000} & \underline{5.867} & \underline{0.157} & \underline{0.895} & \underline{0.974} & \textbf{0.991}\\
                & \textbf{FlowDepth(ours)} & 2 (-1,0) & & 512$\times$192 & \textbf{0.097} & \textbf{0.974} & \textbf{5.693} & \textbf{0.152} & \textbf{0.901} & \textbf{0.975} & \textbf{0.991}\\
                \bottomrule[0.5mm]
            \end{tabular}}
        \end{center}
        \vspace{-0.5cm}
    \end{table*}
    
    \begin{figure*}
        \centering
        \includegraphics[width=\textwidth, trim={0 10 0 0}]{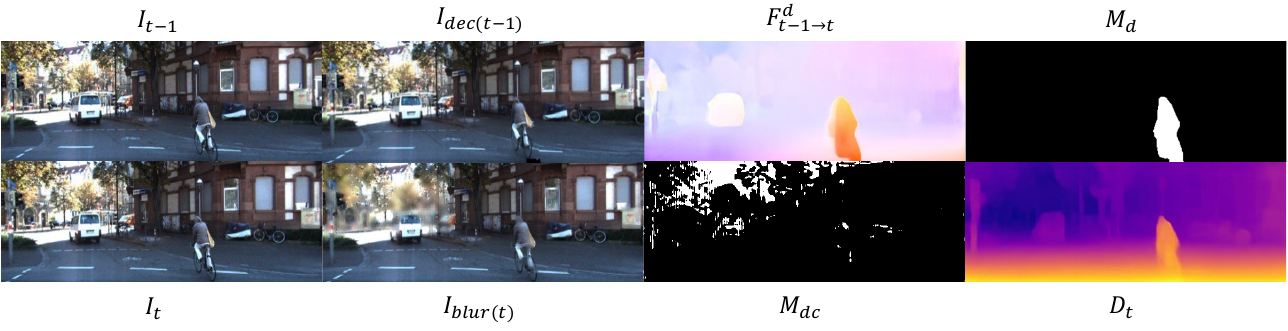}
        \caption{Examples of outputs of each part on KITTI.}
        \label{fig:kitti}
        \vspace{-0.5cm}
    \end{figure*}

    \begin{figure*}
        \centering
        \includegraphics[width=\textwidth, trim={0 15 0 0}]{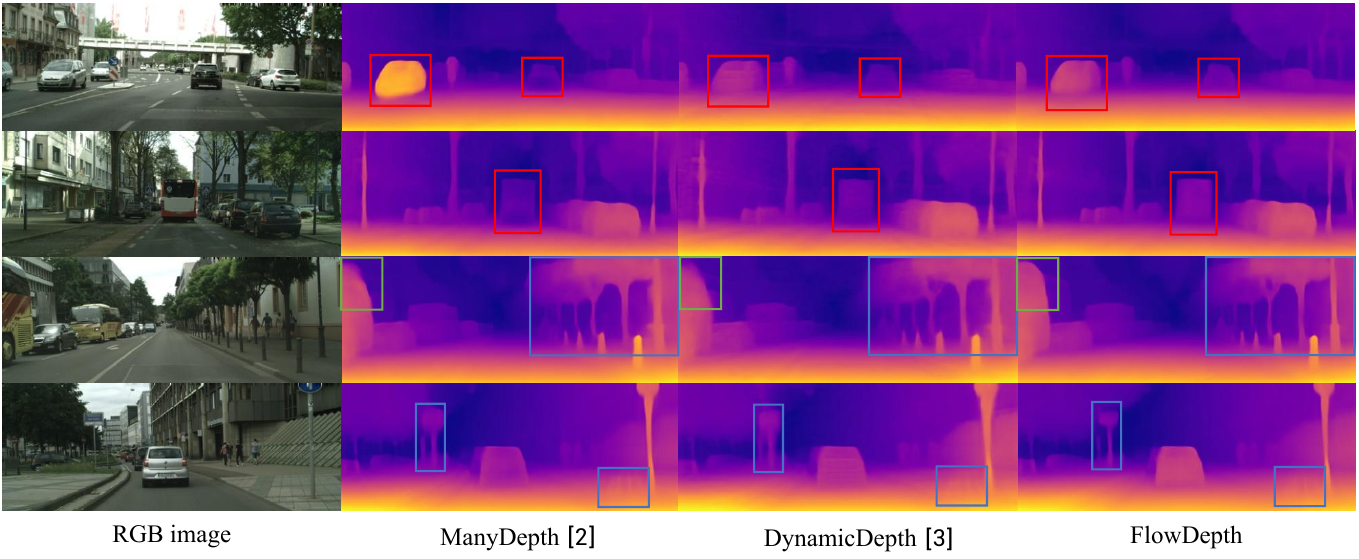}
        \caption{Examples of depth estimations on Cityscapes. Different color boxes represent: \textcolor{red}{Dynamic Objects}, \textcolor[RGB]{64,114,196}{high-freq texture}, and \textcolor[RGB]{112,173,71}{low-texture}}
        \label{fig:cityscapes}
        \vspace{-0.5cm}
    \end{figure*}

    \begin{table*}
        \vspace*{0.25cm}
        \caption{Ablation Study on the Cityscapes Dataset.}
        \label{tab:ablation}
        \normalsize
        \centering
        \resizebox{0.72\textwidth}{!}{
        \begin{tabular}{cccc|cccc|c}
            \toprule[0.5mm]
            \multicolumn{2}{c}{DMFM} & \multirow{2}{*}{DCABlur} & \multirow{2}{*}{CVloss} & \multirow{2}{*}{AbsRel$\downarrow$} & \multirow{2}{*}{SqRel$\downarrow$} & \multirow{2}{*}{RMSE$\downarrow$} & \multirow{2}{*}{RMSE log$\downarrow$} & \multirow{2}{*}{$\delta<1.25 \uparrow$} \\
            w/o mask & w/ mask & & & & & & & \\
            \midrule[0.5mm]
            & & & & 0.113 & 1.191 & 6.217 & 0.168 & 0.879\\
            $\surd$ & & & & 0.111 & 1.219 & 6.132 & 0.165 & 0.883\\
            & $\surd$ & & & 0.101 & 1.019 & 5.851 & 0.157 & 0.896\\
            & $\surd$ & $\surd$ & & 0.100 & 0.997 & 5.720 & 0.154 & \textbf{0.901}\\
            & $\surd$ & & $\surd$ & 0.098 & 0.982 & 5.797 & 0.156 & 0.900\\
            & $\surd$ & $\surd$ & $\surd$ & \textbf{0.097} & \textbf{0.974} & \textbf{5.693} & \textbf{0.152} & \textbf{0.901}\\
            \bottomrule[0.5mm]
        \end{tabular}}
        \vspace{-0.2cm}
    \end{table*}

    \textbf{Network Architecture.} The overall architecture is depicted in Fig.\ref{fig:model}. The depth and camera motion prior networks, and multi-frame depth estimation network are designed following Monodepth2\cite{Godard_Aodha_Firman_Brostow_2018}. with ResNet18 backbones pre-trained on ImageNet. For the optical flow prior network, we employ RAFT-small\cite{Teed_Deng_2020}, which can be self-supervised trained and provide excellent performance for ensuring the effective warping of dynamic regions. In DFMF, the $N_{mask}$ utilizes a U-Net architecture with three layers and a sigmoid activation function. In the DCABlur module, the $E_{s2r}$ and $N_{s2r}$ inside Depth Cue are the same as \cite{Chen_Wang_Chen_Zeng_2021}, while $D_{dc}$ consists of three upsampling layers. \textbf{It is important to emphasize that} all parts of our model only use RGB images to train because both depth estimation and optical flow estimation tasks adhere to geometric consistency thus allowing for self-supervised training.
    
    \textbf{Training.} Before we train the FLowDepth, we will pre-train a depth cue model as described before. We fix $E_{s2r}$ and $N_{s2r}$ and use the output of the prior depth network to supervise the training of $D_{dc}$ for 40 epochs with an Adam optimizer. The initial learning rate is set to 2e-4 and will be decreased by 50\% every 10 epochs. The batch size is set to 8. 
    Then, just like other methods, the entire training process of FlowDepth is divided into two stages. The first stage is to train teacher networks. As most multi-frame methods do, we use frames \{$I_{t-1}$,$I_{t}$,$I_{t+1}$\} for training as well as frames \{$I_{t-1}$,$I_{t}$\} for testing. This stage trains both the prior networks and the multi-frame network for 2 epochs with an Adam optimizer. We set the batch size to 8, and the learning rate to 1e-5.
    The second stage is to fix all prior networks and $N_{mask}$, and only train the multi-frame network with the learning rate of 1e-6 for another 8 epochs.
    The initial value of the learnable parameters are set as $\{\lambda_{cv},\lambda_{d},\lambda_{oof},\lambda_{dof},\lambda_{m}\}=\{0.2,1,1,10,1\}$.
    
    \textbf{Dataset.} To keep consistency with previous work and ensure fairness, we conducted experiments on the KITTI dataset \cite{Menze_Geiger_2015}. Furthermore, as the KITTI dataset contains a small number of dynamic objects, we also report the performance on the more challenging Cityscapes dataset \cite{Cordts_Omran_Ramos_Rehfeld_Enzweiler_Benenson_Franke_Roth_Schiele_2016}, which contains more moving objects. To further validate the transferability of our model, we also do experiments on our own VECAN dataset.
        
\subsection{Results and Analysis} 
    We compare our FlowDepth with prior state-of-the-art methods. The final depth map is capped to 80m and is normalized using median scaling\cite{Godard_Aodha_Brostow_2017}.
    
    \textbf{KITTI results.} The results on the KITTI dataset by the testing split of \cite{Eigen_Fergus_2015} are shown in the upper half of TABLE \ref{tab:results}. As previous works do, we rank all methods based on Frames and AbsRel. Our results demonstrate that FlowDepth surpasses the performance of other comparative methods. In comparison to the baseline, FlowDepth relatively outperforms ManyDepth by 5.1\% in AbsRel. Besides, when compared to the SOTA DynamicDepth, FlowDepth exhibits a relative improvement of 3.1\% in AbsRel, substantiating its effectiveness in multi-frame monocular depth estimation. To ensure a fair comparison, we evaluate our model to TriDepth alone because it directly employs semantic ground truth as input. So we also leverage the semantic ground truth as $M_d$ for the moving objects directly. The new experimental result is shown in the row of FlowDepth$^{*}$. We can find that the model has a significant improvement compared to the original FlowDepth and also outperforms TriDepth$^{*}$ by 2.2\% in AbsRel. This is because the original FlowDepth uses the network to learn the mask, which may filter out distant or slow-moving objects, resulting in these objects not being 'staticized' by the DMFM module. In contrast, using semantic information as a mask can address this issue. A visualization of every part's result in FlowDepth is shown in Fig.\ref{fig:kitti}. However, the number of dynamic objects in KITTI is too small, making the DMFM module proposed much less useful, so we focus on the Cityscapes dataset.

    \begin{table}
        \caption{Depth Error on  Dynamic Objects on the Cityscapes Dataset.}
        \label{tab:dyna_error}
        \normalsize
        \centering
        \resizebox{0.48\textwidth}{!}{
        \begin{tabular}{c|cccc|c}
            \toprule[0.5mm]
            Methods & AbsRel$\downarrow$ & SqRel$\downarrow$ & RMSE$\downarrow$ & \makecell{RMSE\\log$\downarrow$} & \makecell{$\delta$ \\$<1.25 \uparrow$}  \\
            \midrule[0.5mm]
            MonoDepth2 \cite{Godard_Aodha_Firman_Brostow_2018} & 0.159 & 1.944 & 6.461 & 0.214 & 0.820 \\
            ManyDepth \cite{Watson_Mac_Aodha_Prisacariu_Brostow_Firman_2021} & 0.164 & 2.140 & 6.597 & 0.219 & 0.779  \\
            DynamicDepth \cite{Feng_Yang_Jing_Wang_Tian_Li} & 0.129 & 1.274 & 4.626 & 0.168 & 0.862 \\
            FlowDepth &\textbf{ 0.122} & \textbf{1.203} & \textbf{4.561} & \textbf{0.159} & \textbf{0.874} \\
            \bottomrule[0.5mm]
        \end{tabular}
        }
    \end{table}

    \begin{table}
        \caption{Model Complexity Analysis.}
        \label{tab:complexity}
        \normalsize
        \centering
        \resizebox{0.48\textwidth}{!}{
        \begin{tabular}{ccccc}
            \toprule[0.5mm]
            Methods & AbsRel & \makecell{RunTime\\(ms)} & \makecell{Inference \\ params\ (Mb)} & \makecell{Total \\ params\ (Mb)}  \\
            \midrule[0.5mm]
            ManyDepth \cite{Watson_Mac_Aodha_Prisacariu_Brostow_Firman_2021} & 0.114 & 34.0 & 29.8 & 29.8 \\
            \hline
            \rule{0pt}{12pt}
            DynamicDepth \cite{Feng_Yang_Jing_Wang_Tian_Li} & 0.103 & 191.8 & 85.5(+55.7) & 85.5(+55.7)\\
            FlowDepth &  0.097 & 78.5 & 54.2(+24.4) & 79.0(+49.2) \\
            \bottomrule[0.5mm]
        \end{tabular}
        }
        \vspace{-0.5cm}
    \end{table}
        
    \textbf{Cityscapes results.} FlowDepth outperformed other SOTA methods in all metrics, as shown in the bottom half of TABLE \ref{tab:results} on Cityscapes. FlowDepth relatively outperforms ManyDepth by 14.0\% and DynamicDepth by 5.8\% in AbsRel. A comparison example is shown in Fig.\ref{fig:cityscapes}, showcasing a significant improvement in depth estimation. For dynamic objects, there are no more holes and unreasonable depth results, which verifies the DMFM module. In high-freq texture regions, our model not only produces smoother results compared to other methods, as evident in the road in the fourth row, but it also has the ability to distinguish more accurately between foreground objects and backgrounds, such as the trees or the signs. This capability underscores how DCABlur effectively preserves valuable depth cues while blurring useless texture edges during the training stages. In low-texture regions, our model can estimate more precise depths, as demonstrated by the gradient depth estimation of the bus in the third row, This result validates the effectiveness of our proposed cost-volume sparse loss.

    TABLE \ref{tab:dyna_error} shows the depth error on dynamic objects on the Cityscapes dataset. Our proposed DMFM can effectively improve the depth estimation accuracy in dynamic regions, surpassing all previous works on all metrics.
    
    \textbf{Complexity analysis.} Running time and model parameter size are listed in TABLE \ref{tab:complexity}. Our method and DynamicDepth are both based on the ManyDepth framework. However, compared to DynamicDepth which uses the semantic segmentation model EfficientPS\cite{Mohan_Valada_2021}, our model has a smaller size of model parameters both in training and inference, and it has an inference speed that is approximately 2.5 times faster. This is because the proposed DCABlur module and CVloss are only used during the training to help the model find the correct optimization direction. During the inference process, only the optical flow prior network RAFT-small in DMFM is used, which is smaller and faster compared to EfficientPS.

    \begin{table}
        \caption{Model Transferability Analysis.}
        \label{tab:transferability}
        \normalsize
        \centering
        \resizebox{0.48\textwidth}{!}{
        \begin{tabular}{cc|cc|c}
            \toprule[0.5mm]
            Methods & W$\times$H & AbsRel$\downarrow$ & RMSE$\downarrow$ & $\delta<1.25 \uparrow$ \\
            \midrule[0.5mm]
            ManyDepth \cite{Watson_Mac_Aodha_Prisacariu_Brostow_Firman_2021} & 512$\times$192 & 0.116 & 4.779 & 0.871 \\
            DynamicDepth \cite{Feng_Yang_Jing_Wang_Tian_Li} & 512$\times$192 & 0.109 & 4.686 & 0.883 \\
            FlowDepth &  512$\times$192 & \textbf{0.096} & \textbf{4.243} & \textbf{0.899} \\
            \bottomrule[0.5mm]
        \end{tabular}
        }
        \vspace{-0.5cm}
    \end{table}
    
    \begin{figure*}
        \vspace*{0.25cm}
        \centering
        \setlength{\abovecaptionskip}{0.cm}
        \includegraphics[width=\textwidth]{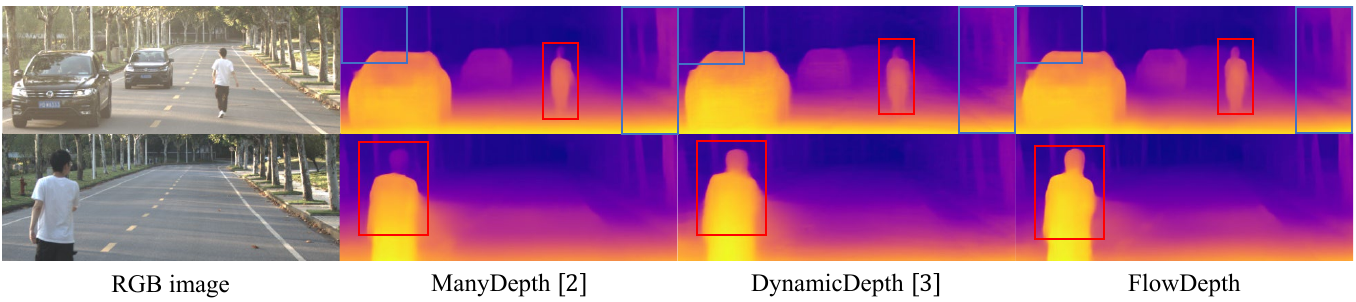}
        \caption{Examples of depth estimations on our own VECAN Dataset which is used to verify the model transferability.}
        \label{fig:vecan}
        \vspace{-0.6cm}
    \end{figure*}

    \textbf{Transferability analysis.} We also further tested the transferability of FlowDepth on VECAN datasets collected by 'RUIYU' autonomous driving platform, independently developed by the VECAN lab at Tongji University, to validate its practical application capabilities. TABLE \ref{tab:transferability} presents a comparison of quantitative metrics, while Fig.\ref{fig:vecan} shows a comparison of qualitative results. Since FlowDepth does not rely on any additional labeling and only requires video data for training, it demonstrates superior transferability compared to DynamicDepth, which can only use a frozen semantic segmentation network.

\subsection{Ablation Study}
    In order to further analyse the role of each module proposed, we conducted ablation experiments on Cityscapes. As shown in TABLE \ref{tab:ablation}, we examined the effects of DMFM, DCABlur and CVLoss respectively. \textbf{DMFM:} Comparing the first three rows, we can see that adding the DMFM module without $N_{mask}$ improves the performance slightly, but there is a larger decline for the SqRel metric due to the fact that the $F^d$, which is directly calculated by prior, contains the warp of background. Therefore, the introduction of $N_{mask}$ solves this problem and improves the overall performance. \textbf{DCABlur:} Comparing the third and fourth rows, we can find that DCABlur mainly works on the high-freq regions to correct the fusion problem of foreground and background, which get a boost in the absolute metrics (RMSE/RMSE log). \textbf{CVLoss:} From the third and fifth rows, we can see that CVLoss does help on low-texture regions, so it has more improvement for the relative evaluation metrics (AbsRel/SqRel). Therefore, each module contributes to the overall depth estimation, and when all modules work together, they can significantly improve the accuracy.

\section{Conclusion}
\label{sec:conclu}

    We proposed a novel self-supervised multi-frame monocular depth estimation model FlowDepth. It decouples the moving objects using depth, pose, and flow prior in order to solve the mismatch problem caused by dynamic objects. The depth-cue-aware blur operation and the cost-volume loss also mitigate the unfairness of reprojection loss due to the high-freq or low-texture regions in the images. With the proposed innovations, our model outperforms other methods on the KITTI and Cityscapes datasets. Moreover, FlowDepth also demonstrates lower complexity and better transferability.

\bibliographystyle{unsrt}
\bibliography{root}

\end{document}